\documentclass[11pt,a4paper]{article}
\usepackage{authblk}
\usepackage[hyperref]{acl2021}
\usepackage{times}
\usepackage{latexsym}
\usepackage{booktabs,chemformula}

\usepackage{stackengine}
\usepackage{amsmath}
\usepackage{multirow}
\usepackage{url}
\usepackage{float}
\usepackage{pgfplots}
\usepackage{footnote}
\usepackage[flushleft]{threeparttable}
\usepackage[utf8]{inputenc}
\usepackage[T5]{fontenc}
\usepackage{longtable}
\usepackage{supertabular}
\usepackage{makecell}
\usepackage{multicol}
\usepackage{array}
\usepackage{lipsum}
\usepackage{placeins}

\usepackage{flafter}

\usepackage{pifont}%

\usepackage{microtype}

\usepackage{graphicx}
\graphicspath{ {./images/} }

\aclfinalcopy 

\DeclareUnicodeCharacter{0301}{\'{e}}
\newcommand{\footref}[1]{%
    $^{\ref{#1}}$%
}
\newcolumntype{L}[1]{>{\raggedright\let\newline\\\arraybackslash\hspace{0pt}}m{#1}}
\newcolumntype{C}[1]{>{\centering\let\newline\\\arraybackslash\hspace{0pt}}m{#1}}
\newcolumntype{R}[1]{>{\raggedleft\let\newline\\\arraybackslash\hspace{0pt}}m{#1}}

\makeatletter
\@ifpackageloaded{hyperref}{%
  \renewcommand\footref[1]{%
    \begingroup 
    \unrestored@protected@xdef\@thefnmark{%
      \ref*{#1}%
    }%
    \endgroup 
    \ifHy@hyperfootnotes 
       \expandafter\@firstoftwo 
    \else 
       \expandafter\@secondoftwo 
    \fi 
    {\hyperref[#1]{\strut\H@@footnotemark}}{\@footnotemark}%
  }%
}{}%

\newcommand\savedlabel{}%
\AtBeginDocument{\let\savedlabel=\label}%
\newcommand\footnotereflabel[1]{%
   \@bsphack
   \begingroup
   \def\@currentHref{Hfootnote.\theHfootnote}\savedlabel{#1}%
   \endgroup
   \@esphack
}%

\makeatother

\title{UIT-E10dot3 at SemEval-2021 Task 5: Toxic Spans Detection with Named Entity Recognition and Question-Answering Approaches}

\author[1, 2]{\textbf{Phu Gia Hoang}}
\author[1, 2]{\textbf{Luan Thanh Nguyen}}
\author[1, 2]{\textbf{Kiet Van Nguyen}}
\affil[1]{University of Information Technology, Ho Chi Minh City, Vietnam}
\affil[2]{Vietnam National University Ho Chi Minh City, Vietnam}
\affil[ ]{{\ttfamily\{19520215, 17520721\}@gm.uit.edu.vn, kietnv@uit.edu.vn}}

\date{}
  
\date{}
\begin{document}
\maketitle
\begin{abstract}

The increment of toxic comments on online space is causing tremendous effects on other vulnerable users. For this reason, considerable efforts are made to deal with this, and SemEval-2021 Task 5: Toxic Spans Detection is one of those. This task asks competitors to extract spans that have toxicity from the given texts, and we have done several analyses to understand its structure before doing experiments. We solve this task by two approaches, Named Entity Recognition with spaCy’s library and Question-Answering with RoBERTa combining with ToxicBERT, and the former gains the highest F1-score of \textbf{66.99\%}.

\end{abstract}

\section{Introduction}

The world of social media is overgrowing, and users easily express their opinions or feelings toward topics that they are concerned about. However, because of the freedom of speech, lots of toxic comments or contents are uncontrollably increasing. There are several kinds of research about the effect of toxic speech on users' health. In 2017, research about the impact of toxic language on health was conducted \cite{mohan2017impact}. Sometimes, with toxic words, conversations can become cyberbullying, cyber threats, or online harassment, which are harmful to users. To reduce those negative impacts, there are abundant researches for classifying contents into toxic or non-toxic, and then they hide the whole text if it is toxic. However, that action may inhibit the freedom of speech. As a result, censoring only toxic spans is the better solution for this problem. Therefore, in SemEval-2021 Task 5: Toxic Spans Detection \cite{pav2020semeval} we try to realize it.

About toxic contents on the internet, researches were only about binary toxicity classification. Still, in task 5 of SemEval-2021, which is about toxic spans detection, we conduct more in-depth research into the toxicity, find exactly which parts of the text are toxic. As the NER approach and Question-Answering (QA) approach, we propose two approaches for solving this problem. We use RoBERTa \cite{liu2019roberta} combining with ToxicBERT \cite{Detoxify}, transfer learning models, for QA approach and spaCy's library \cite{honnibal2017spacy} for NER approach.

We organize the paper as follows. Section 2 is related works that we consult for building the systems. The dataset and analyses are defined in Section 3. In section 4, we introduce our two proposed systems for toxic spans detection. Section 5 describes the results of the studies and analyses. Finally, in Section 6, we bring our work to a close.
\section{Related Works}
Researchers around the world these days have started to concentrate on toxic speech. It inflicts individual and group harm, damaging our social fabric \cite{tirrell2018toxic}. Several datasets for classifying toxicity on toxic speech on online forums, such as the dataset provided by \citet{waseem2016hateful} for English, BEEP! dataset for Korean by \citet{moon2020beep}, the dataset for Russian provided by \citet{Smetanin2020Toxic}, TolD-Br dataset for Brazilian Portuguese by \citet{leite2020toxic}, and UIT-ViCTSD, a dataset about constructive and toxic speech detection for Vietnamese by \citet{nguyen2021constructive}.

Besides, there are shared tasks about toxic speech as well as hate speech such as these from SemEval, includes SemEval-2019 Task 5 Multilingual Detection of Hate \cite{basile2019semeval}, SemEval-2019 Task 6 Identifying and Categorizing Offensive Language in Social Media (OffensEval) \cite{zampieri2019semeval}, SemEval-2020 Task 12 Multilingual Offensive Language Identification in Social Media \cite{zampieri2020semeval2020}, and SemeEval-2021 Task 5 Toxic Spans Detection \cite{pav2020semeval}, which is the current task we have to deal with in this paper.
\section{Dataset}
The origin of this SemEval-2021 Task 5 dataset comes from the publicly available Civil Comments dataset \cite{borkan2019nuanced}, which consists of 1.2M posts and comments. The data in this public dataset have no annotation of any toxic spans in toxic posts but do have post-level toxicity annotations, which mean showing which posts or entire of them are toxic. And the holders of this task retain 30K of them, which were annotated to be toxic or severely toxic by at least half of the crowd-raters from annotations of Borkan et al.

The task holders then randomly keep 10K posts from the 30K posts for annotating toxic spans. They employ three experienced crowd-raters per post from a third-party crowd-annotation platform, and they warn them about adult content. However, task organizers also claim that not all toxic posts are annotated with toxic spans.

The task for crowd-raters is to highlight toxic sequences of the comments, and if the comment is not toxic or should annotate the whole of it, crowd-raters have to check the appropriate box without highlighting any spans. Consequently, we have two columns, the spans column and the text column. The spans column has lists of numbers or null that reference toxic character offsets in the text column, and some of the given data are shown in the following table.

\begin{table}[H]
\caption{Examples for the given data.}
\small
\begin{tabular}{p{3.45cm}|p{3.45cm}}
\hline
\textbf{spans}                                                                                                        & \textbf{text}                                                                                                \\ \hline
{[}7, 8, 9, 10, 11, 12{]}                                                                                             & Pretty \textbf{damned} eloquent ... :)                                                                                \\ \hline
{[}0, 1, 2, 3, 4, 5, 6, 7, 8, 9, 10, 11, 12, 13, 14, 15, 16, 17, 18, 19, 20, 21, 22, 23, 24, 25, 26, 27, 28{]}        & \textbf{He might fire you to the moon}, but you already have a head full of cheese!                                   \\ \hline
{[}0, 1, 2, 3, 4, 5, 6, 7, 8, 9, 15, 16, 17, 18, 19, 20, 21, 22, 23, 24, 98, 99, 100, 101, 102, 103, 104, 105, 106{]} & \textbf{Nauseating} and \textbf{disgusting}. Thank goodness the First Amendment permits people to demonstrate their \textbf{stupidity}. \\ \hline
{[}{]}                                                                                                                & Not if they shoot you first...                                                                               \\ \hline
\end{tabular}
\label{tab:my-table}
\end{table}

The competitors receive two separate training and test sets from organizers. In the training set, there are 7,939 records, and in the test set, there are 2,000 records. Furthermore, as mentioned in the data annotating process, one text that possibly has multiple toxic spans is highlighted. Figure 1 and Figure 2 illustrate the distribution of spans in the training and the test sets.

\begin{figure}[H]
    \centering
    \includegraphics[width=1\linewidth]{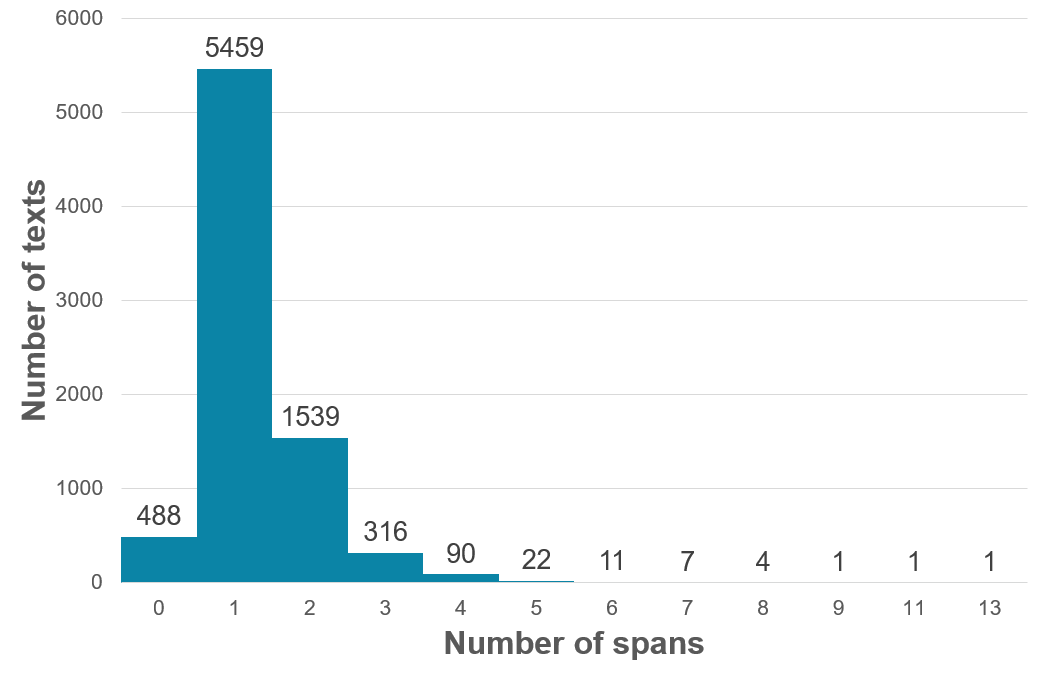}
    \caption{Distribution of spans in the training set.}
    \label{fig:data_train_dist}
\end{figure}

\begin{figure}[H]
    \centering
    \includegraphics[width=1\linewidth]{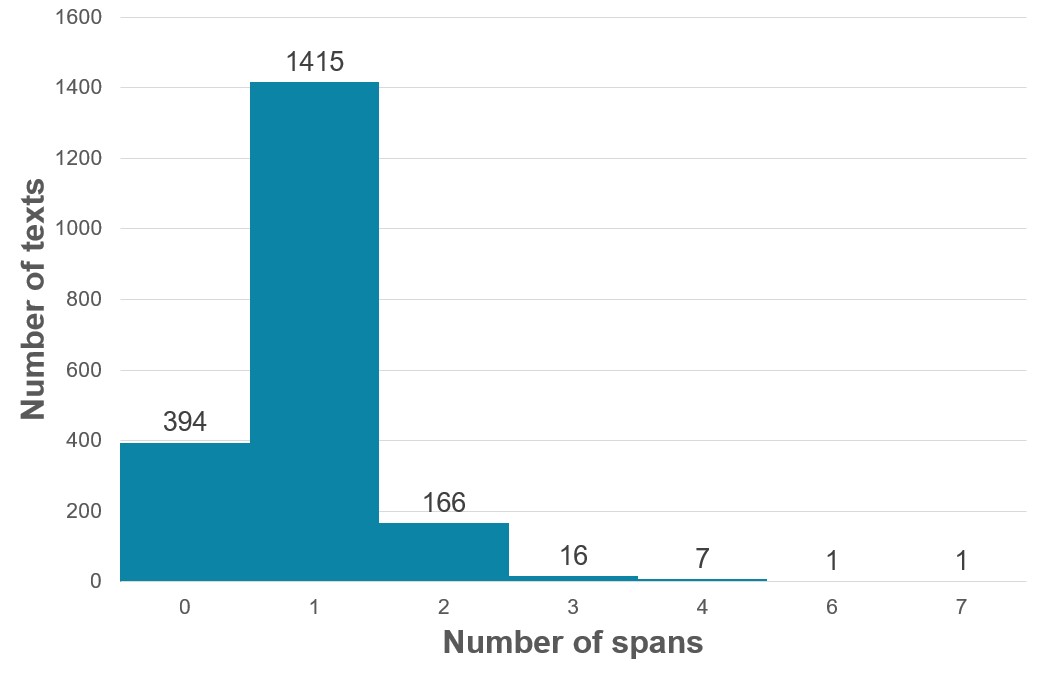}
    \caption{Distribution of spans in the test set.}
    \label{fig:data_test_dist}
\end{figure}

For more details, according to Figure \ref{fig:data_train_dist} and Figure \ref{fig:data_test_dist}, there is a significant number of single spans in each post, and it accounts for nearly 68.8\% and 70.8\% in the training set and the test set, respectively. It is also interesting to notice that the number of zero spans is not tiny, and the proportion of it in the training set is less than in the test set, more specifically, 19.7\% in the test set and 6.15\% in the training set.

Moreover, we also calculated the Jaccard score of text and spans in the given dataset for more in-depth analysis. The Jaccard score, also known as the Jaccard index or Jaccard similarity coefficient, was developed by Paul Jaccard \cite{jaccard1912distribution} and it is a statistic used for measuring the similarity and diversity of sample sets as follows.


\begin{figure}[H]
    \centering
    \includegraphics[width=1\linewidth]{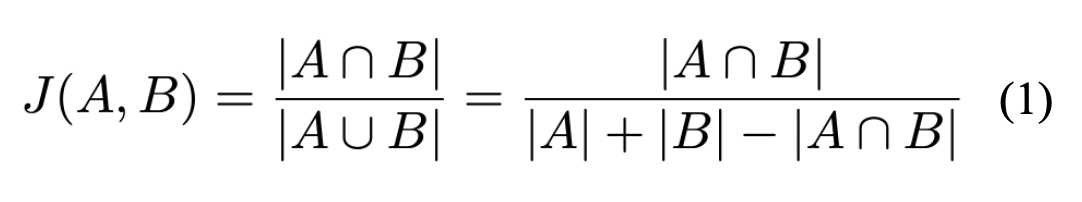}
\end{figure}

\begin{figure}[H]
    \centering
    \includegraphics[width=1\linewidth]{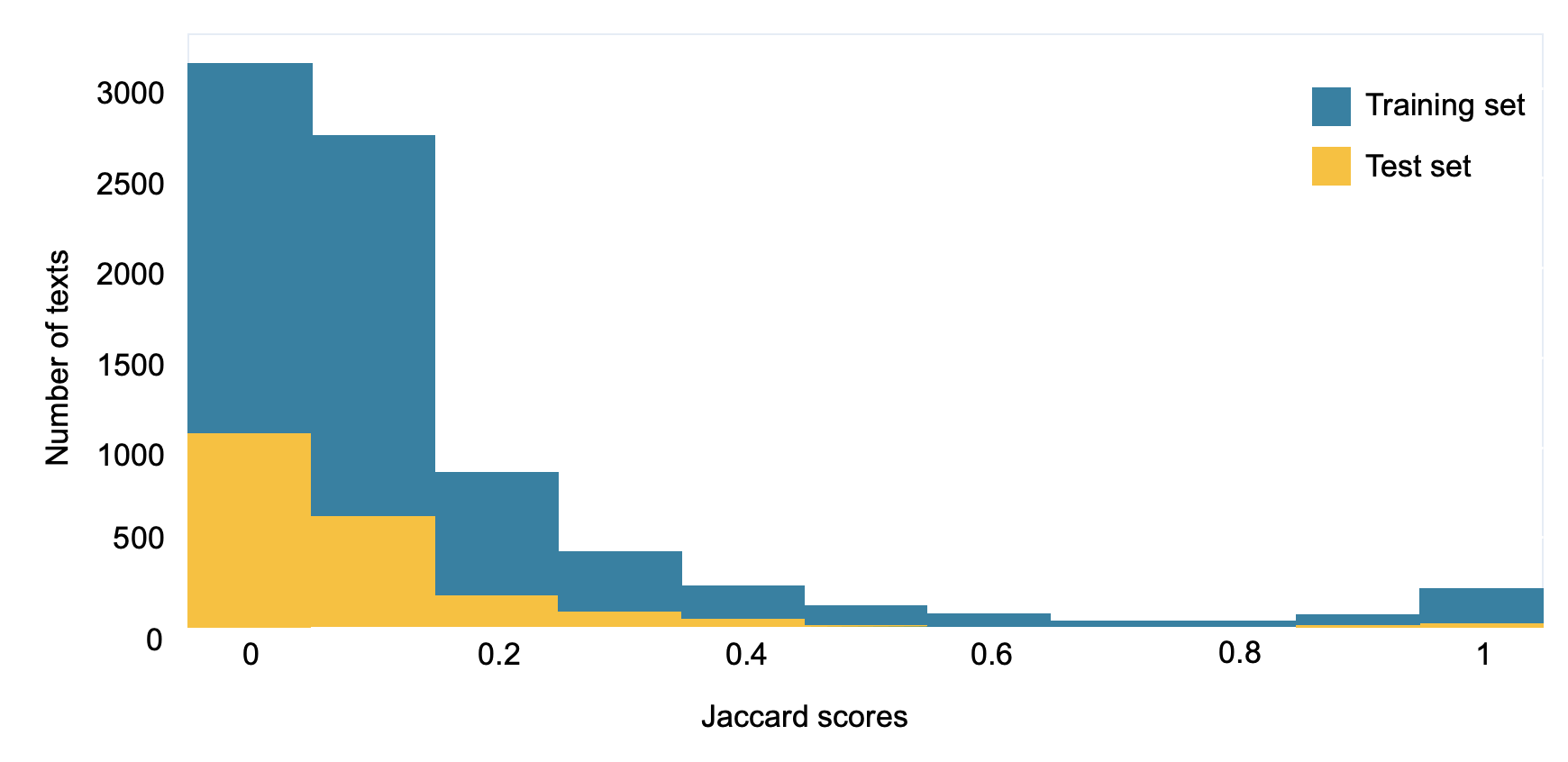}
    \caption{Histogram of Jaccard score of each record in the training set and the test set.}
    \label{fig:Jaccard_scores}
\end{figure}

The histogram in Figure \ref{fig:Jaccard_scores} illustrates that most of the data points have Jaccard scores in the range of 0 to 0.35, and the peak is at 0 to 0.05, which means toxic character offsets are just a fraction in each post even there are records annotated all characters of the post are toxic. There are 16 records in the test set and 212 records in the training set with Jaccard scores at 0.95 to 1.0. For that reason, just the toxic part(s) of the comments needs to be censored rather than the whole comment as in the traditional method.
\section{Systems}
In this paper, we propose two systems for the toxic spans detection task with NER and QA approaches. The first system is the QA approach based on RoBERTa and the second system is the NER approach based on spaCy's library.

\subsection{Question-Answering Approach Based on RoBERTa}
\begin{figure}[H]
    \centering
    \includegraphics[width=1.\linewidth]{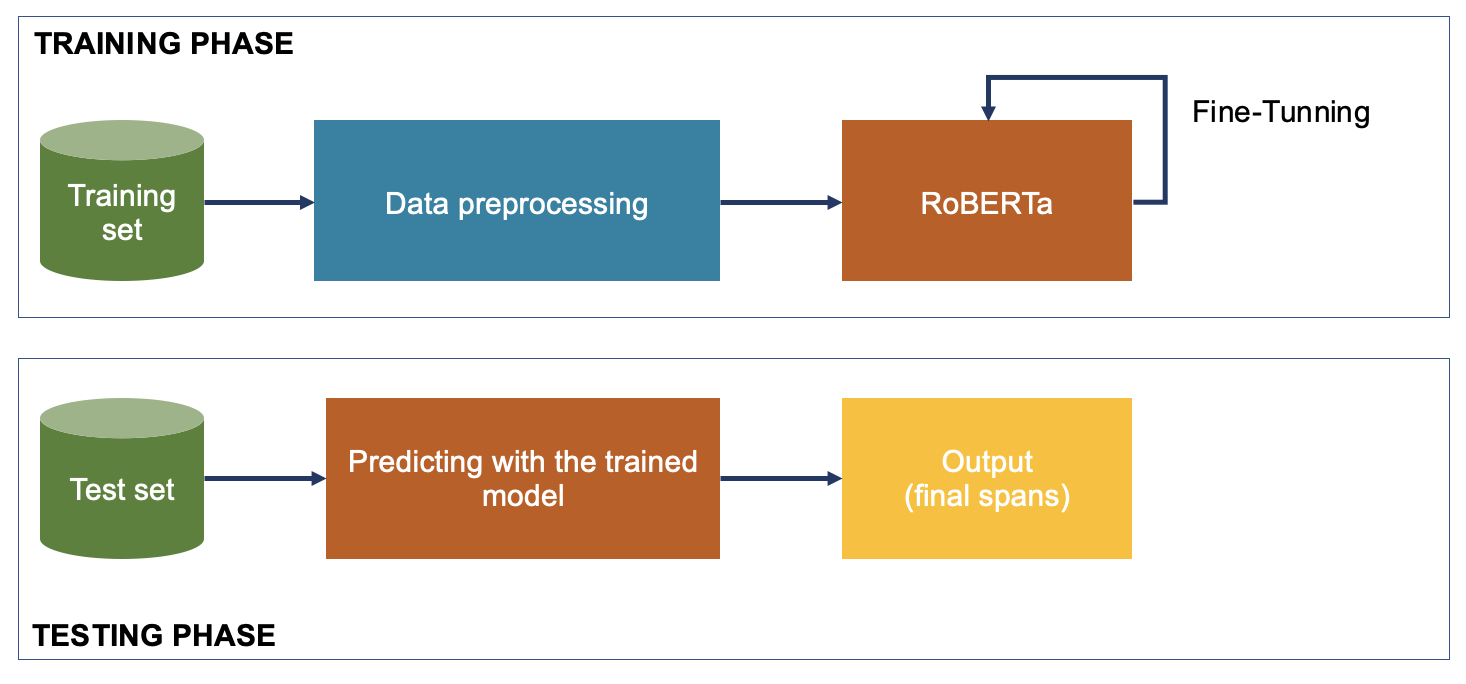}
    \caption{Training and testing phases of toxic spans detection with RoBERTa based system.}
    \label{fig:RoBERTa_base_system}
\end{figure}

With the QA approach, we use RoBERTa combining with ToxicBERT as the basis for the system. RoBERTa \cite{liu2019roberta} is a transfer learning model and it is a replication study of BERT \cite{devlin2019bert}. Unlike BERT, to improve the training performance, RoBERTa eliminates the Next Sentence Prediction (NSP) task of the pre-trained model BERT. ToxicBERT \cite{Detoxify} is also a transfer learning model, and it uses BERT as the main model for classifying toxicity. ToxicBERT has an outstanding performance for the task of Jigsaw Unintended Bias in Toxicity Classification \footnote{https://www.kaggle.com/c/jigsaw-unintended-bias-in-toxicity-classification} on Kaggle, which uses the same dataset with SemEval-2021 Task 5, with 93.64\% F1-score. We use two models for our QA approach system, and the overview of the system with training and testing phases is described in Figure \ref{fig:RoBERTa_base_system}.

\begin{figure}[H]
    \centering
    \includegraphics[width=1.\linewidth]{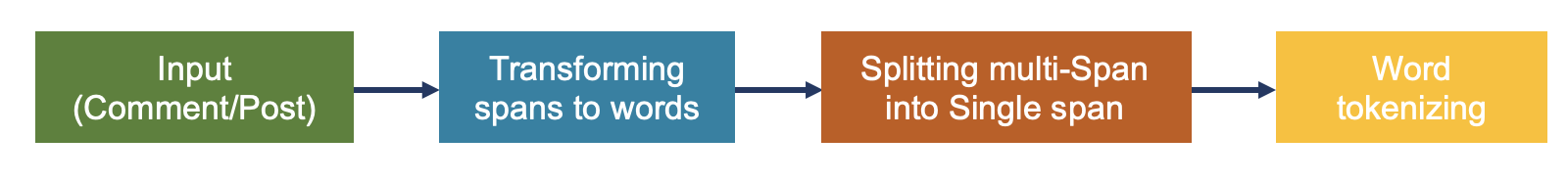}
    \caption{Data preprocessing of RoBERTa based system for toxic spans detection.}
    \label{fig:RoBERa_Datapreprocessing}
\end{figure}

\begin{figure*}[]
    \centering
    \includegraphics[width=1.\linewidth]{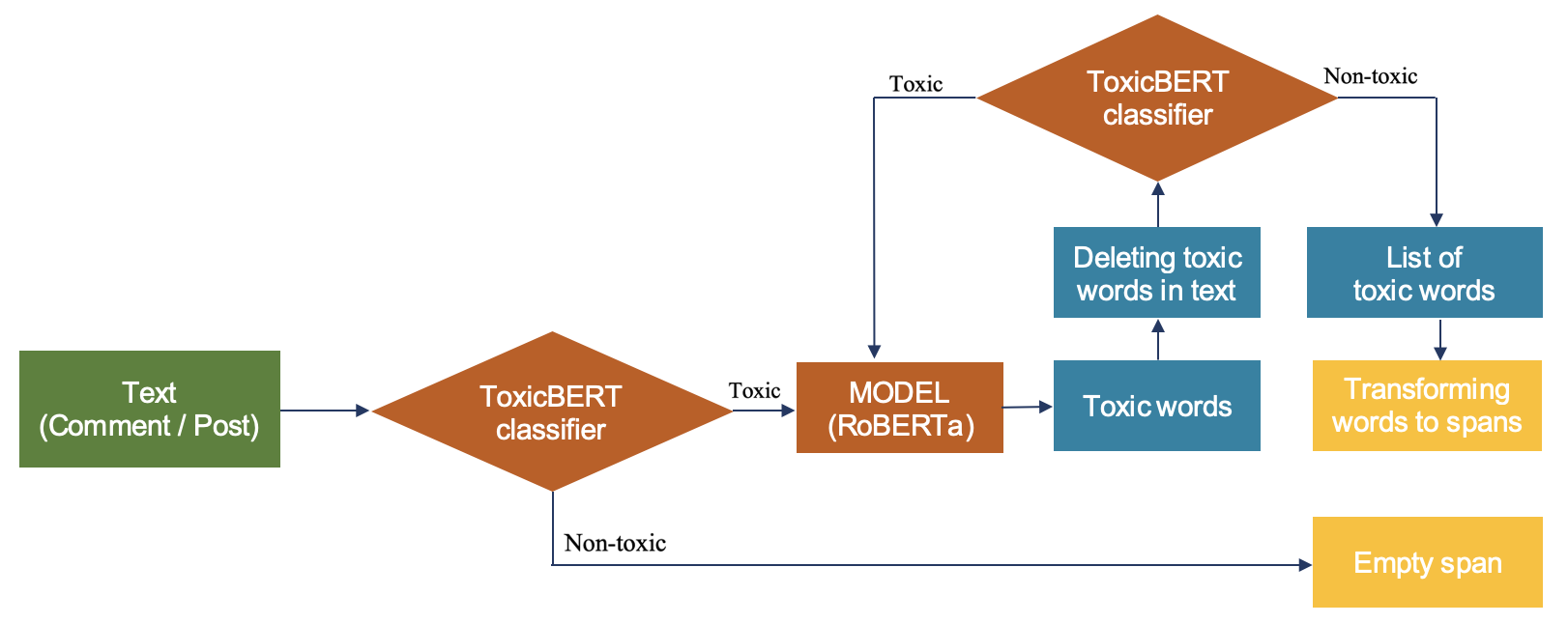}
    \caption{Predicting toxic spans with trained model by QA approach.}
    \label{fig:RoBERa_Predicting}
\end{figure*}

Firstly, we preprocess the training set with techniques to get the right format for the RoBERTa model, mentioned in Figure \ref{fig:RoBERa_Datapreprocessing}. The model we used only approve one spans, but several examples have more than one in the training set, and we called it "multi-span". Hence, we split multi-span (*) into single spans (**) (***) as below.

\begin{itemize}
    \item Plain text: 
    
    (*) This bitch is so fucking idiot.
    
    \item After splitting: 
    
    (**) This \textbf{bitch} is so.
    
    (***) This is so \textbf{fucking idiot}.
\end{itemize}

After splitting texts, we tokenize the dataset with a subword model as Byte-Pair Encoding (BPE) \cite{sennrich2015neural}. Then, we feed the data into a pre-trained RoBERTa model and fine-tune it with suitable parameters. We analyze the length of the texts in the dataset and set max\_length=512 and epochs=5 for the model. After searching for extensive hyper-parameters, we set the learning\_rate and drop\_out equal to 3e-5 and 0.1, respectively. We also train the model with 5-fold cross-validation. After the training phase, the trained RoBERTa model is used for predicting new toxic spans.

In the testing phase, besides using RoBERTa, we use another transfer learning model is ToxicBERT \cite{Detoxify} for identifying toxic comments. With ToxicBERT, we classify the input text into toxic or non-toxic labels before predicting spans. If the result is non-toxic, we stop the prediction, and the result is an empty spans. If it is toxic, we feed the text into the RoBERTa model to predict toxic words. After having the spans, to ensure that the text still has toxic words, we remove the predicted toxic word(s) from the processing text and then recheck its toxicity by ToxicBERT and re-predict its remaining toxic words (if any). 

Because final results are words, we transform them into spans for the requirement of this task. 

\subsection{NER Approach Based on spaCy's Library}
In this approach, we tag all the characters spans with text as TOXIC to train the model, and we predict all TOXIC tags in the text set of texts.

For solving this, we choose version 2.2.5 of spaCy's NER Model \cite{honnibal2017spacy} because of its exceptionally efficient statistical system in both speed and accuracy for this named-entity recognition. Apart from default entities such as location, person, organization, and so on, spaCy also enables training the model with new entities by updating it with newer examples.

\begin{figure}[H]
    \centering
    \includegraphics[width=1.\linewidth]{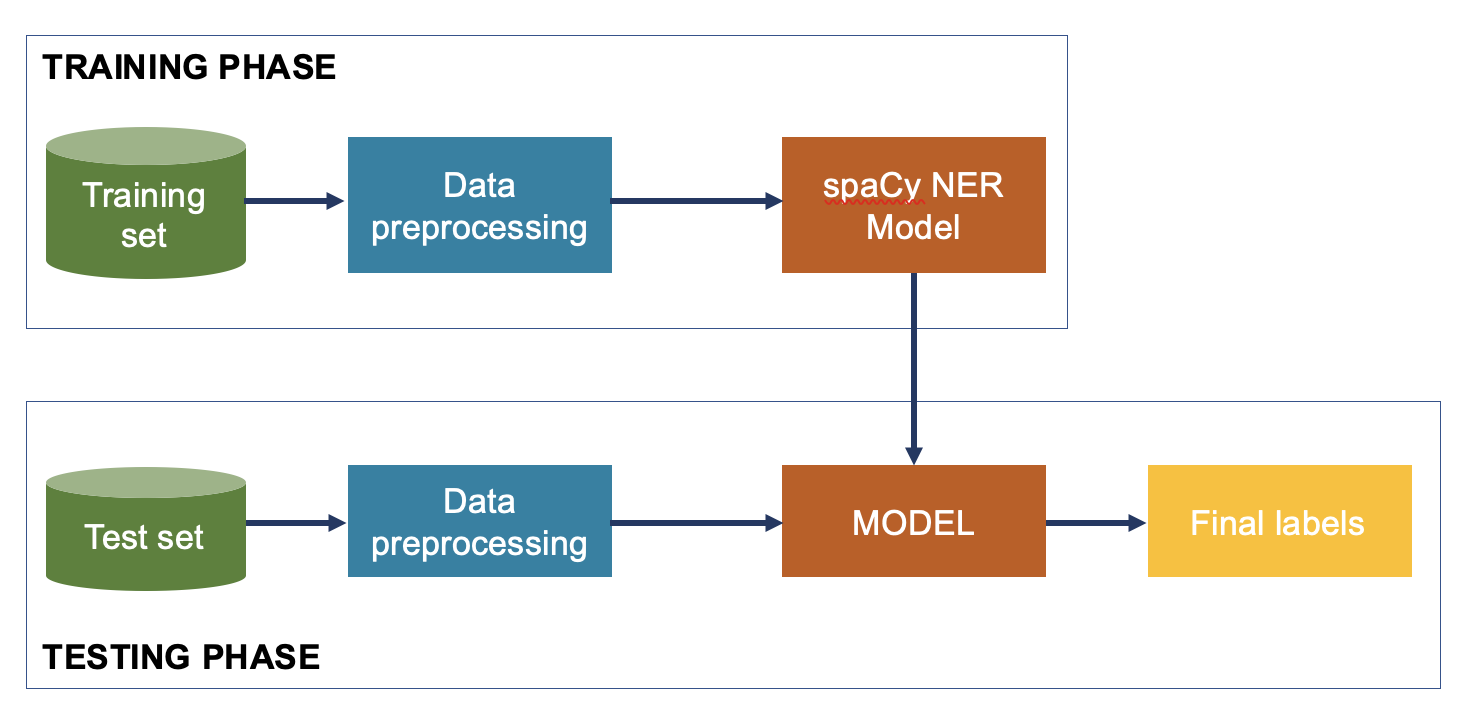}
    \caption{Training and testing phases of toxic spans detection with spaCy based system.}
    \label{fig:spaCy_Overview}
\end{figure}

The above Figure \ref{fig:spaCy_Overview} shows the process of our spaCy based system. Both training and test sets have to be tokenized before feeding them into the spaCy NER model or being predicted by the TOXIC entities. For more details, in the training phase, the input data have to be in the right format for the spaCy NER model as in the following Figure \ref{fig:spaCy_format}.

\begin{figure}[H]
    \centering
    \includegraphics[width=1.\linewidth]{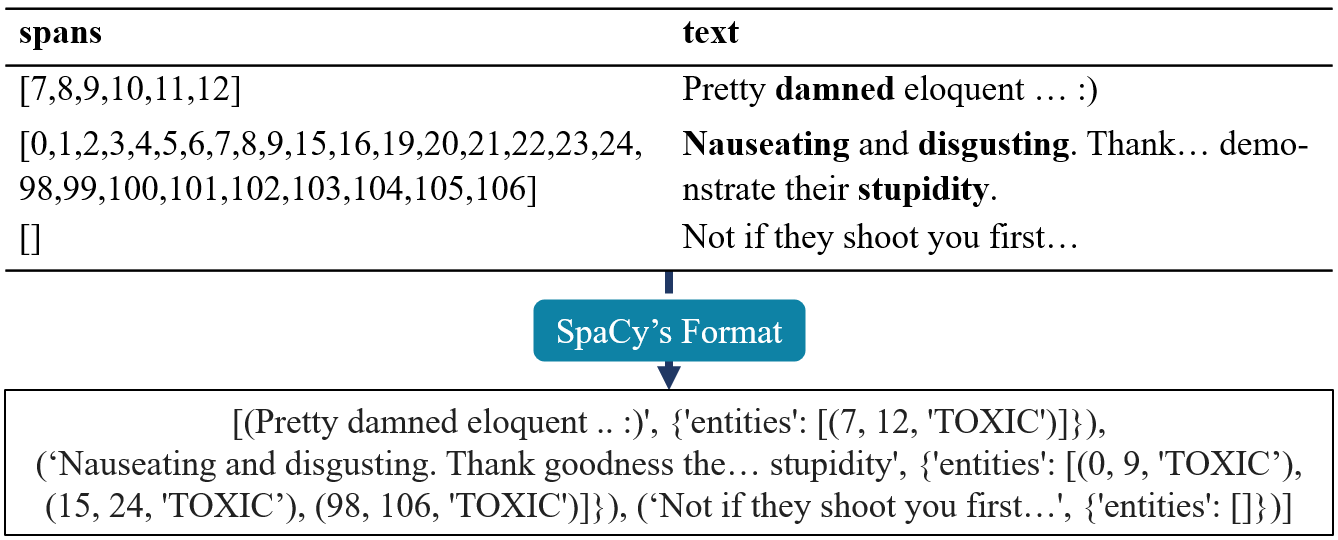}
    \caption{Process of re-formatting data for spaCy based system.}
    \label{fig:spaCy_format}
\end{figure}

SpaCy has not published the architecture of their models yet, but they do have a brief explanation about how their models work, especially the NER model, through a four-step formula: embed, encode, attend, and predict.

\begin{figure*}[]
    \centering
    \includegraphics[width=1.\linewidth]{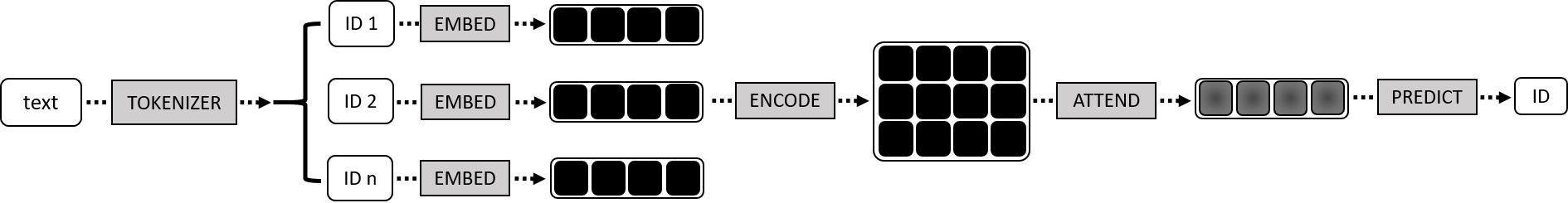}
    \caption{Diagram of how spaCy's models work.}
    \label{fig:spaCy_base_system}
\end{figure*}

As in Figure \ref{fig:spaCy_base_system}, spaCy's model is fed with unique numerical values (ID) which address a token of a corpus or a class of the NLP task (named entity class). In the first embed stage, word similarities are revealed by extracting hash, which is collected by extracting word features as the lower case, the prefix, the suffix, and the shape. The encode stage is fed with a sequence of word vectors from the previous stage to calculate a representation which is named sentence matrix. In the sentence matrix, the meaning of each token in the context of neighboring tokens is represented in each row, and this is done by using a bidirectional RNN \cite{schuster1997bidirectional}. The output matrix from the second stage is injected into the Attention Layer of the CNN after summarized by a query of vectors. Finally, to predict the toxic class, a softmax function is utilized. After the model is trained, the CNN model is now used for the NER task to extract the toxic class.

The given toxic spans dataset is fed into spaCy’s library for training with a suitable format. During the contest, my team was using spaCy's library for a small model for English (en\_core\_web\_sm) at version 2.2.5, and we tried different parameters to get the optimal result. When training, the dataset is shuffled and passed through spaCy’s training algorithm in batches with an increment of batch sizes from 4.0 to 32.0 and step of 1.001. Moreover, the drop rate is consistently at 0.5, and most of the experiments loop 45 times.
\section{Experiments}

After building two such systems, we start to experiment on the test set, and the following subsections dicuss our results.

\subsection{Evaluation Metrics}
Before going through experimental results, we first discuss the evaluation metrics used in this SemEval-2021 Task 5. 

In this task, all of the responding systems from participants are evaluated by \textbf{F1} score \cite{da2019fine}. Assuming the system \textbf{$S_{i}$} returns \textbf{$C^{t}_{S_{i}}$}, which is a toxic character offsets of the post. Let \textbf{$G^{t}$} be the character offsets of the ground truth annotation of \textbf{t}. In the following formulas, the \textbf{F1} score of system \textbf{$S_{i}$} is computed regarding ground truth \textbf{G} of post \textbf{t} ($\left | \cdot \right |$ indicates set cardinality).

\begin{figure}[H]
    \centering
    \includegraphics[width=0.9\linewidth]{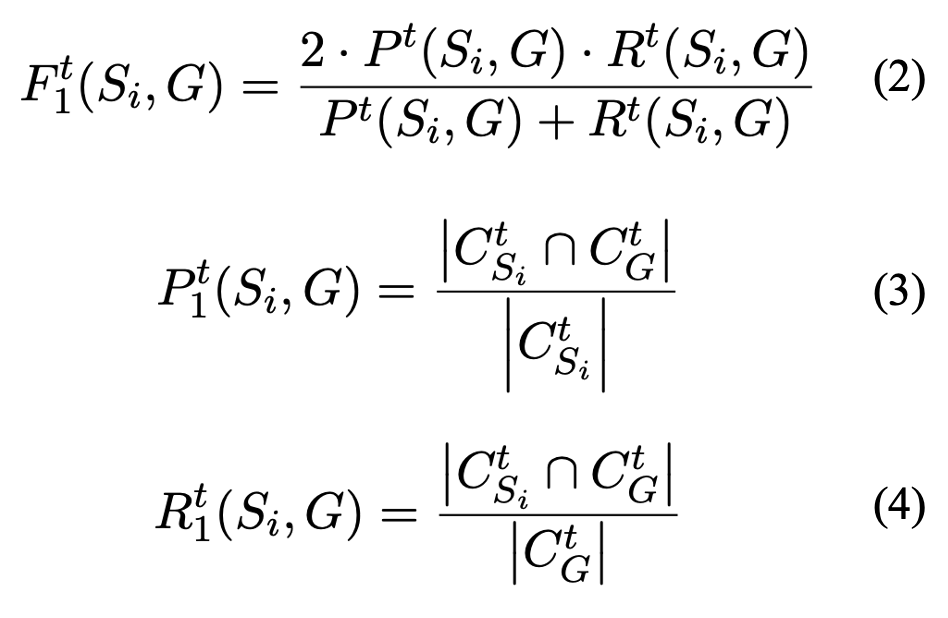}
\end{figure}

If \textbf{$S^{t}_{G}$} is empty for posts \textbf{t}, we set \textbf{$F^{t}_{1}(S_{i},G)$} = 1 and if \textbf{$S^{t}_{A_{i}}$} is empty, \textbf{$F^{t}_{1}(S_{i},G)$} = 0. Finally, we calculate average of \textbf{$F^{t}_{1}(S_{i},G)$} of all over the posts \textbf{t} if test set to get a sigle \textbf{F1} score of the system \textbf{$S_{i}$}.

\subsection{Experimental Results}
The results of our systems compared with other teams' are shown in Table \ref{tab:results}.

\begin{table}[H]
\caption{The results of our systems compared with other teams by F1-score (\%).}
\label{tab:results}
\begin{tabular}{llll}
\hline
\textbf{Rank}                & \textbf{Team name}                    & \textbf{}           & \textbf{F1-score} \\ \hline
1                            & HITSZ-HLT                             &                     & 70.83            \\ \hline
2                            & S-NLP                                 &                     & 70.77            \\ \hline
33                           & lz1904                                &                     & 67.00            \\ \hline
\multirow{2}{*}{\textbf{34}} & \multirow{2}{*}{\textbf{UIT-E10dot3}} & \textbf{spaCy} & \textbf{66.99}   \\ \cline{3-4} 
                             &                                       & RoBERTa             &   52.12                \\ \hline
\multicolumn{3}{l}{}                                                                       & 17.00±1.       \\ \hline
\end{tabular}
\end{table}

During the SemEval-2021 Task 5, with the spaCy base system, we achieved rank 34 out of 91 teams, and in the table above, we have shown our result with the spaCy based system and the RoBERTa based system in comparison with rank 1, 2, 33 and random baseline of this task. The F1-score of our best system is 66.99\%, 3.84\% lower than the first rank team, and 49.09\% higher than the baseline model.

\subsection{Result Analyses}
After analyzing our most effective system based on spaCy's library, we spot crucial errors in predicting and datasets by comparing predicted spans to gold spans. Several records in the given data are standing alone without the context that leads to confusing or multi-meaning. Moreover, comments are using slang(s) or idiom(s), causing null output for our system. We also realize a lack of consistency or highlighting non-toxic spans when annotating data about the datasets. Likewise, several words in the text have spelling mistakes that intentionally also impair our system performance. Evidence for those errors are in Table \ref{tab:results_analysis}, Appendix.
\section{Conclusion and Future Work}
In this paper, we introduced two proposed systems for toxic spans detection based on named entity and question-answering approaches. We obtained the highest results with the SpaCy's library based system with the F1-score of \textbf{66.99\%} and ranked 34 out of 91 teams in SemEval-2021 Task 5. 

In future, we plan to improve our systems by implementing various SOTA models for toxic spans detection. With the built systems, we can create friendly online conversations and make social media forums safer for users.

\section*{Acknowledgement}
We would like to express our gratitude for the helpful reviews from reviewers.

\bibliographystyle{acl_natbib} 
\bibliography{anthology, UIT-E10dot3} 

\appendix

\begin{table*}[]
\caption{Examples for result analyses.}
\label{tab:results_analysis}
\begin{tabular}{@{}p{1cm}p{7cm}p{7cm}@{}}
\toprule
\textbf{No.} & \textbf{Predicted by spaCy base system}                                                                                                                                                                                                                          & \textbf{Ground Truth}                                                                                                                                                                                                                                            \\ \midrule
\multicolumn{3}{c}{\textit{Requiring context (\ding{171})}}                                                                                                                                                                                                                                                                                                                                                                                                                                                                                                      \\ \midrule
1            & No, my poor benighted correspondent, your reductionism reveals an archaic mindset. Intersex, questioning, nongendered, and a vast variety of others have rights too!                                                                                             & No, my poor \textbf{benighted} correspondent, your reductionism reveals an archaic mindset. Intersex, questioning, nongendered, and a vast variety of others have rights too!                                                                                             \\ \midrule
\multicolumn{3}{c}{\textit{Slang(s) / Idiom(s)}}                                                                                                                                                                                                                                                                                                                                                                                                                                                                                                  \\ \midrule
2            & Lord have Mercy on us, Trump is running amok.                                                                                                                                                                                                                    & Lord have Mercy on us, Trump is running \textbf{amok}.                                                                                                                                                                                                                    \\
3            & No, dumdum-you dont get it. The business does not get to decide who he serves and who he does not. If he makes cakes for one, he must make them for ALL                                                                                                          & No, \textbf{dumdum}-you dont get it. The business does not get to decide who he serves and who he does not. If he makes cakes for one, he must make them for ALL                                                                                                          \\
4            & Ugh, the worst, burn them at the stake.                                                                                                                                                                                                                          & Ugh, the worst, \textbf{burn them at the stake}.                                                                                                                                                                                                                          \\ \midrule
\multicolumn{3}{c}{\textit{Lacking of consistency or highlighting non-toxic spans when annotating (\ding{168})}}                                                                                                                                                                                                                                                                                                                                                                                                                                                                                  \\ \midrule
5            & There is even a website where sore loser Democrats can pretend that Hillary won. La-la-land indeed! \#liberaltears \#salt \#schadenfreude                                                                                                                        & There is even a website where sore \textbf{loser} Democrats can pretend that Hillary won. La-la-land indeed! \#liberaltears \#salt \#schadenfreude                                                                                                                        \\
6            & Hey Repubes, I bet you think that "wall" is going to be built on Mexicos dollar too huh? What a bunch of gullible \textbf{losers}!!!!                                                                                                                                     & Hey Repubes, I bet you think that "wall" is going to be built on Mexicos dollar too huh? What a bunch of gullible losers!!!!                                                                                                                                     \\
7            & \textless{}Trump was a draft dodger during Vietnam.\textgreater The president is a coward and a hypocrite who not only dodged the draft and avoided any and every opportunity to serve… No one has more contempt for the flag than does Donald J. Trump. No one. & \textless{}Trump was a draft dodger during \textbf{Vietnam.\textgreater} The president is a coward and a hypocrite who not only dodged the draft and avoided any and every opportunity to serve… No one has more contempt for the flag than does Donald J. Trump. No one. \\ \midrule
\multicolumn{3}{c}{\textit{Spelling mistake}}                                                                                                                                                                                                                                                                                                                                                                                                                                                                                                     \\ \midrule
8            & FUKC YOU @QuinHillyer YOU ARE THE ONE NEED TO GO, YOU ANTI AMERICA FAKE MOTHER FUKCER                                                                                                                                                                            & \textbf{FUKC YOU} @QuinHillyer YOU ARE THE ONE NEED TO GO, YOU ANTI AMERICA FAKE \textbf{MOTHER FUKCER}                                                                                                                                                                            \\
9            & Hey Repubes, I bet you think that "wall" is going to be built on Mexicos dollar too huh? What a bunch of gullible \textbf{losers}!!!!                                                                                                                                     & Hey Repubes, I bet you think that "wall" is going to be built on Mexicos dollar too huh? What a bunch of gullible losers!!!!                                                                                                                                     \\ \bottomrule
\multicolumn{3}{p{16cm}}{ }\\
\multicolumn{3}{p{16cm}}{(\ding{171}) The table shows that with this example, our system predicts that there are no toxic spans; meanwhile, in ground truth, the word \textbf{benighted} is highlighted to be toxic. We assume that because of lacking the context of the text and the word \textbf{benighted} also has multi-meaning.}                                                       \\
\multicolumn{3}{p{16cm}}{ }\\

\multicolumn{3}{p{16cm}}{(\ding{168}) The table shows that in example No. 5, the word \textbf{loser} is annotated to be toxic when in example No. 6, also having the plural form of word \textbf{loser} but not to be highlighted. Meanwhile, in example No. 7, the spans \textbf{Vietnam.\textgreater{}} is highlighted even if it does not have toxicity.}                                                                                
\end{tabular}
\end{table*}



\end{document}